\documentclass[journal]{IEEEtran}

\usepackage{cite}
\usepackage{amsmath,amssymb,amsfonts}
\usepackage{graphicx}
\usepackage{multirow}
\usepackage{booktabs}
\usepackage{stfloats}% allows [b] placement of figure*/table*
\usepackage{xcolor}
\usepackage{textcomp}
\usepackage{url}
\usepackage[hidelinks]{hyperref}

\newcommand{\tabhead}[1]{\begin{tabular}[c]{@{}c@{}}#1\end{tabular}}% multi-line table header

\begin{document}

\title{BEVPredFormer: Spatio-Temporal Attention for BEV Instance Prediction in Autonomous Driving}

\author{Miguel~Antunes-García, Santiago~Montiel-Marín, Fabio~Sánchez-García, Rodrigo~Gutiérrez-Moreno, Rafael~Barea, and~Luis~M.~Bergasa%
\thanks{The authors are with the Electronics Department, University of Alcalá (UAH), 28805 Alcalá de Henares, Madrid, Spain (e-mail: miguel.antunes@uah.es; santiago.montiel@uah.es; fabio.sanchezg@uah.es; rodrigo.gutierrez@uah.es; rafael.barea@uah.es; luism.bergasa@uah.es). Corresponding author: Miguel Antunes-García.}%
\thanks{This work was supported by project PID2024-161576OB-I00, funded by the Spanish MICIU/AEI/10.13039/501100011033 and co-funded by the European Regional Development Fund (ERDF, ``A way of making Europe''), and by project PLEC2023-010343 (INARTRANS 4.0), funded by MICIU/AEI/10.13039/501100011033, and Spanish MICIU through a FPU grant (FPU23/02905). The authors acknowledge support from iRoboCity2030-CM project (grant TEC-2024/TEC-62), awarded by the Community of Madrid, and the support of the ELLIS Unit Madrid.}}

\maketitle

\begin{abstract}
A robust awareness of how dynamic scenes evolve is essential for Autonomous Driving systems, as they must accurately detect, track, and predict the behaviour of surrounding obstacles.
Traditional perception pipelines that rely on modular architectures tend to suffer from cumulative errors and latency.
Instance Prediction models provide a unified solution, performing Bird's-Eye-View segmentation and motion estimation across current and future frames using information directly obtained from different sensors.
However, a key challenge in these models lies in the effective processing of the dense spatial and temporal information inherent in dynamic driving environments.
This level of complexity demands architectures capable of capturing fine-grained motion patterns and long-range dependencies.
We introduce BEVPredFormer, a novel multi-camera architecture for Bird's-Eye-View Instance Prediction that uses attention-based temporal processing to improve temporal and spatial comprehension of the scene and relies on an attention-based 3D projection of the camera information.
BEVPredFormer employs a recurrent-free design that incorporates gated transformer layers, divided spatio-temporal attention mechanisms, and multi-scale prediction heads.
Additionally, we incorporate a difference-guided feature extraction module that enhances temporal representations.
Extensive ablation studies validate the design of the temporal module, including the configuration of the spatio-temporal attention blocks and the difference module, as well as the choice of feature extractor and view transform, the training strategy and the input resolution.
When evaluated on the nuScenes dataset, BEVPredFormer performs on par with or surpasses state-of-the-art methods, particularly in long-range scenarios, at a moderate computational cost.
Code is available at \url{https://github.com/miguelag99/BEVPredFormer}.
\end{abstract}


\section{Introduction}\label{sec1}

\IEEEPARstart{A}{} robust understanding of the scene surrounding the ego vehicle is essential for any advanced Autonomous Driving (AD) system, which involves diverse goals such as estimating the position and predicting the intentions of other dynamic actors in the scene.
Traditionally, perception architectures have addressed these tasks using a series of dedicated modules performing complementary functions, such as 3D object detection, multi-object tracking, and future motion prediction.
Although effective, this multi-stage approach can introduce cumulative errors and additional processing latencies.
By contrast, Instance Prediction models \cite{fiery2021} offer a more comprehensive solution by performing Bird's-Eye-View (BEV) segmentation and motion estimation for current and future frames simultaneously.
This approach provides a unified BEV map showing the positions and movements of different instances, obtained directly from raw sensor data. This integrated approach simplifies the complex perception pipeline and inherently captures the spatio-temporal context.

AD Instance Prediction methods often rely on data from multiple sensors, such as cameras, Light Detection and Ranging (LiDAR), and Radio Detection and Ranging (RADAR), to capture a detailed view of the surrounding environment of the vehicle.
When using image information, image features are first projected from the image space into a BEV representation \cite{philion2020liftsplat,li2022bevformer}. This representation provides a structured and spatially consistent format for subsequent sensor fusion and processing.
To capture how the scene evolves over time, these models process sequences of past frames, requiring a temporal module to model motion and temporal dependencies.
Several approaches have been explored to model temporal information. These include 2D and 3D convolution-based methods \cite{powerbev,fiery2021}, which efficiently capture spatio-temporal features across sequences of frames. Recurrent Neural Networks (RNNs), such as Long Short-Term Memory (LSTM) and Gated Recurrent Unit (GRU) networks \cite{stretchbev}, provide another approach by sequentially processing past frames and capturing long-range temporal dependencies.
Attention-based modules \cite{detracasas} further enable models to dynamically focus on relevant regions or features across different frames, improving their ability to handle complex temporal relationships effectively.
Finally, the output is obtained using dedicated prediction heads. The specific design and output of these heads can vary depending on the method.
While simple designs such as PowerBEV \cite{powerbev} use only segmentation and flow outputs, other models, such as Fiery \cite{fiery2021}, include additional heads to predict centerness or offsets for more precise object-level processing.

The work discussed in this paper introduces BEVPredFormer, a state-of-the-art (SOTA) architecture designed for Instance Prediction in the BEV space.
By incorporating attention-based temporal processing, BEVPredFormer captures temporal dependencies across past frames, improving its ability to model the evolution of the surrounding scene.
Unlike recurrent-based approaches such as StretchBEV \cite{stretchbev}, our architecture adopts a recurrent-free design in the temporal processing module in order to improve parallelisation and computational efficiency.
Based on PredFormer \cite{tang2024predformer}, the proposed method uses gated transformer layers to process BEV temporal information through divided spatial and temporal attention mechanisms.
PredFormer leverages attention mechanisms to perform video prediction on sequences of images, alternating and combining spatial and temporal attention layers \cite{timesformer} to capture the dynamic scene evolution.
BEVPredFormer adapts this concept by modifying the attention encoder to operate on a sequence of BEV feature maps extracted using a BEVFormer-based BEV projection block.
This adaptation enables the model to learn temporal dependencies and spatial relationships effectively in the BEV domain, thereby enhancing its ability to predict future states in AD scenarios.
Following PredFormer, gated transformer layers are used to selectively control the flow of information within the temporal module.
To further enhance the temporal representation, we incorporate a module designed to extract difference features \cite{dmp} between the BEV features extracted from consecutive input frames.
To enable accurate future prediction, a multi-scale prediction module \cite{powerbev,dmp} projects temporally enriched features into future time steps.

While BEVPredFormer builds on some well-established BEV perception and prediction modules, the main novelty of this work lies in the adaptation of divided spatio-temporal attention, originally designed for video frame prediction, to sequences of multi-camera BEV features for Instance Prediction, and in the analysis of how its configuration affects performance in this domain.

The main contributions of this study are summarised below:
\begin{itemize}
    \item BEVPredFormer, a new multi-camera BEV Instance Prediction architecture that combines an attention-based BEV projection with a recurrent-free, attention-based temporal module, is proposed.
    \item A temporal processing block that adapts PredFormer from video frame prediction to BEV feature sequences, using difference-guided features as input and gated transformer layers with divided temporal and spatial attention, whose configuration (type, order and depth) is systematically analysed.
    \item Extensive ablation studies are performed on the temporal module, the feature extractor and view transform, the training strategy and the input resolution to assess their influence on the overall performance.
    \item Performance evaluation and SOTA comparison are performed on the nuScenes AD dataset, achieving SOTA performance in most metrics.
\end{itemize}

The rest of the paper is organised as follows.
Section~\ref{sec2} reviews related work on multi-camera BEV perception, vehicle future prediction and camera temporal processing.
Section~\ref{sec3} describes the proposed architecture, including the feature extraction, BEV projection, temporal module and prediction heads.
Section~\ref{sec4} presents the experimental setup, the evaluation metrics, the comparison with SOTA methods, the ablation studies and a robustness analysis.
Finally, Section~\ref{sec5} summarises the conclusions of this work.

\section{Related Work}\label{sec2}

\subsection{Multi-Camera BEV Perception}\label{subsec2.1}

Multi-camera setups are frequently employed in AD systems due to their lower cost compared to other sensors such as LiDAR or RADAR, and their ability to provide complete coverage of the vehicle's surroundings with high semantic density.
Consequently, architectures that focus on the transformation of the information present in the different camera planes into a common BEV representation are crucial to achieve a more complete understanding of the scene.
A common BEV representation has a computational complexity advantage over other fully 3D approaches, such as those usually employed for dense tasks such as 3D Occupancy Prediction. Additionally, it facilitates the fusion with features from other sensors if available.
The camera BEV projection methods can be divided into different groups depending on how they project the information: depth-based or Forward Projection, and attention-based or Backward Projection.

Depth-based methods rely on estimating depth maps for each camera to perform the corresponding feature projection.
This methodology was introduced in Lift-Splat-Shoot (LSS) \cite{philion2020liftsplat}, which predicts a categorical depth distribution for each feature position and projects the features into a frustum-shaped region in the final BEV space.
BEVDet \cite{huang2021bevdet} builds on this concept, aiming to build a simpler modular BEV segmentation architecture. The authors also focus on addressing the overfitting problem by introducing data augmentation techniques in the BEV space rather than only on the image plane.
BEVDepth \cite{li2022bevdepth} focuses on reducing depth estimation errors to enhance the camera BEV representation, utilising an additional module for estimating the depth values and incorporating a LiDAR sensor to perform additional supervision during training.
To further enhance scene context understanding, BEVDet4D \cite{huang2022bevdet4d} incorporates temporal processing into its architecture. Initially, multiple past and present BEV feature maps are generated, followed by a spatio-temporal alignment to ensure effective information fusion.
This approach is crucial in Instance Prediction architectures, in which the temporal information is essential to generate the future motion data for each vehicle.

On the other hand, attention-based projection methods rely on attention mechanisms to establish correspondence between the BEV space and the different camera feature maps.
BEVFormer \cite{li2022bevformer} introduces a BEV encoder layer that pulls information from the multi-camera system into the BEV queries using a cross-attention module. A BEV query self-attention block is responsible for further processing temporal information.
BEVFormerV2 \cite{Yang2022BEVFormerVA} focuses more on temporal processing, maintaining the cross-attention module to project the image features and replacing the temporal self-attention block with a temporal encoder that fuses multiple past BEV feature maps.
Both BEVFormer and BEVFormerV2 try to alleviate the computational burden of the attention mechanism using Deformable Attention \cite{defformable}.
To further reduce the complexity of the transformer layers, other approaches employ a sparse approach, where instead of processing every BEV position, only the important ones are processed.
SparseBEV \cite{sparsebev} introduces Scale-Adaptive self-attention to encode the different BEV queries with different receptive fields and an Adaptive Mixing block to efficiently mix and process relevant BEV features.
PointBEV \cite{chambon2024pointbev} proposes a sparse architecture with a two-step inference process.
The first step involves a coarse inference, which identifies and processes relevant features in the scene.
Subsequently, a second fine pass is responsible for processing only the areas surrounding the selected anchor points, enhancing the performance and results.
Approaches such as FB-BEV or FB-OCC \cite{li2023fbbev,li2023fbocc} employ both projection techniques, performing a forward-backward transformation to achieve a better BEV representation.
HENet \cite{henet} introduces a dual-stream encoder that processes two types of temporal inputs. Firstly, a long-term sequence at lower resolution is processed to provide a broader temporal context. Secondly, a short-term sequence with higher-resolution images and a deeper encoder is utilised to capture fine spatial details.
More recently, GaussianCaR \cite{gaussiancar} uses Gaussian Splatting as a view transformer to lift camera and RADAR features into a common sparse 3D representation, which is then projected into BEV, offering an efficient alternative to both projection families.

Different SOTA evaluations suggest that attention-based methods generally outperform depth-based approaches, especially when high-resolution input images are available, since they facilitate more accurate feature extraction and spatial reasoning.
However, hybrid methods that combine depth and attention mechanisms can introduce significant computational overhead, impacting performance.
In this context, BEVPredFormer offers a more streamlined approach by utilising the attention-based projection mechanism of BEVFormer, thus bypassing the need for depth estimation modules.
This design choice also enables direct compensation for ego-motion, as the model generates the 3D grid query positions while taking into account the ego-vehicle movement.

\subsection{Vehicle Future Prediction}\label{subsec2.2}

\subsubsection{Motion Prediction}\label{subsubsec2.2.1}

Conventional SOTA motion prediction pipelines adopt a multi-stage approach, initially detecting objects in the scene to identify the different agents, followed by tracking to associate these agents across frames.
Once tracked, the past trajectories of each agent are used as input to a prediction module that forecasts their future positions, in metres, over a short time horizon.
CRAT-Pred \cite{cratpred} relies on a Graph Neural Network (GNN)-based architecture to process and model interactions between different actors within a scene.
Similarly, HEAT \cite{heat} incorporates High-Definition (HD) map information into the graph.
Recent methods employ attention mechanisms to capture the interactions between the agents.
Attention-GAN \cite{cghnet} first uses a self-attention module to process the social interactions, and then utilises a Generative Adversarial Network (GAN) module to generate the different future trajectories.
VectorNet \cite{vectornet} directly incorporates vectorised information from both scene context and agent behaviour, constructing an attention-based interaction graph.
Gómez-Huélamo et al. \cite{gomez2023efficient} propose an efficient baseline that encodes the social information through a GNN and attention module, also incorporating map information.
More recent transformer-based methods further refine how temporal information is modelled. FINet \cite{finet} incorporates potential future trajectories into the interaction modelling between agents, while DONUT \cite{donut} adopts a decoder-only design that predicts trajectories autoregressively, processing past and future steps with a unified model.

Other proposed architectures address the task of motion prediction using an end-to-end, fully differentiable design, where the latent state of one stage is passed as the input to the next one.
This enables the system to be trained jointly, improving the flow of information across modules.
For example, End2End \cite{e2e_iteraction} performs motion prediction in a two-stage setup, where the detection features and states are processed in the second stage.
UniAD \cite{hu2023_uniad} extends this idea to build a complete AD system that integrates detection, tracking, and prediction in a unified framework. DeTra \cite{detracasas} combines LiDAR and map data in a single-stage model to directly generate future trajectory proposals without requiring intermediate steps.

\subsubsection{Instance Prediction}\label{subsubsec2.2.2}

In contrast to motion prediction, Instance Prediction focuses on forecasting the future state of the full scene in BEV format, often including segmentation, occupancy masks, and motion fields.
This BEV representation divides the scene into a fixed grid that includes information about the state, instance ID and motion for each cell.
These approaches predict the evolution of every instance in the scene directly from the sensor information, being particularly useful for dense forecasting and planning.
Fiery \cite{fiery2021} laid the foundations of Instance Prediction methods, using a multi-camera sensor setup to generate BEV segmentation maps with a probabilistic approach.
BEVerse \cite{beverse} uses a common spatio-temporal BEV encoder to process past information, relying on multiple heads to perform various tasks, including Instance Prediction.
StretchBEV \cite{stretchbev} introduces a stochastic latent-space update approach that processes predictions in space and time, enhancing the diversity and accuracy of longer-term BEV forecasts.
PowerBEV \cite{powerbev} simplifies the process of predicting instances by generating BEV segmentation and flow outputs directly, without the need for additional outputs such as centerness or offset heads.
Fast and Efficient (FaE) \cite{fast_effic} follows this simplified approach, focusing on reducing the computational cost and model complexity of the temporal processing and specialised prediction heads.
DMP \cite{dmp} enhances the temporal representation by introducing a difference-guided motion module.
A related line of work forecasts the future 3D occupancy of the scene instead of BEV instances. For example, IR-WM \cite{irwm} jointly forecasts the future occupancy and plans the ego trajectory using an implicit residual world model. These approaches address a more general but also more computationally demanding 3D formulation, whereas BEV Instance Prediction provides a compact representation for vehicle-level forecasting.

Our proposed architecture performs future Instance Prediction directly from raw sensor data in BEV format, enabling dense and structured scene forecasting.
It captures the full dynamics of the environment by predicting segmentation, occupancy and motion fields for each grid cell, without relying on intermediate representations as required by a motion prediction multi-stage architecture.

\subsection{Camera Temporal Processing}\label{subsec2.3}

\begin{figure*}[t]
    \centering
    \includegraphics[width=0.97\linewidth]{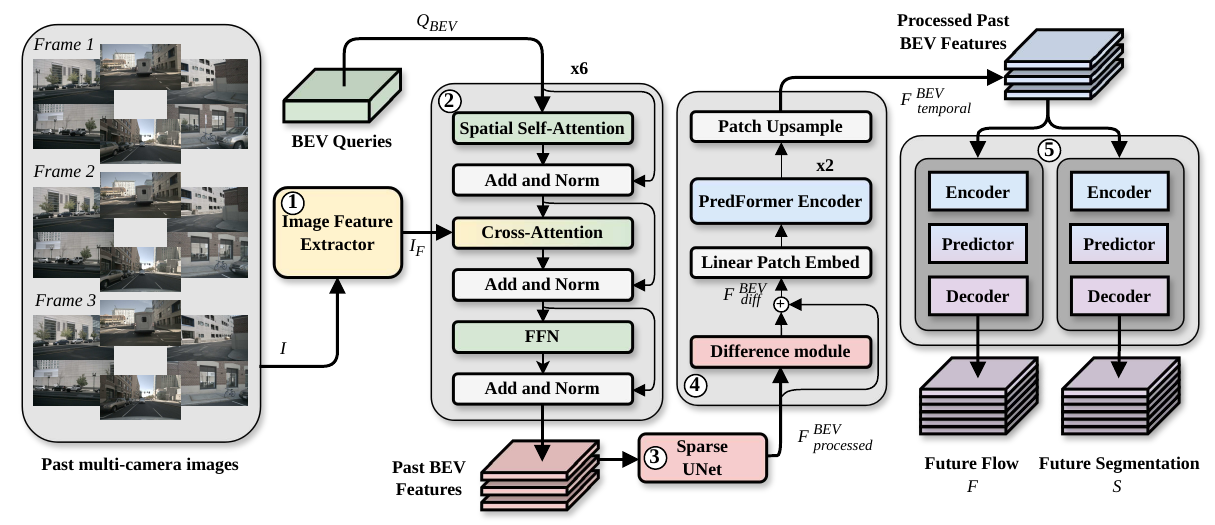}
    \caption{BEVPredFormer proposed architecture}
    \label{fig:architecture}
\end{figure*}

Temporal processing is essential for modelling motion and dynamic changes across frames in camera-based perception.
These methods improve the consistency and accuracy of scene understanding over time by leveraging information from past inputs.
Existing approaches can be categorised as either recurrent or recurrent-free, with different trade-offs in terms of efficiency and performance.

Recurrent approaches such as VMamba \cite{liu2024vmambavisualstatespace} and SwinLSTM \cite{tang2023swinlstm} model temporal dependencies by processing sequences of features frame by frame, maintaining a hidden state across time.
VMamba introduces efficient sequence modelling using Mamba layers, while SwinLSTM combines the spatial modelling power of Swin Transformers with recurrent temporal updates for better video understanding.
In the BEV domain, OnlineBEV \cite{onlinebev} recurrently fuses BEV features over time, aligning past and current features through motion-guided attention.

To improve parallelisation and performance, recurrent-free approaches process all input frames concurrently.
TimeSformer \cite{timesformer} applies factorised space-time attention to video sequences, improving spatio-temporal comprehension and reducing computational costs.
DMVFN \cite{DMVFN} uses a lightweight routing module to predict future frames.
This module adaptively selects motion-specific subnetworks, enabling multi-scale voxel flow modelling with reduced computation while maintaining high accuracy.
PredFormer \cite{tang2024predformer} introduces Gated Transformer Blocks (GTBs) to enhance temporal learning by modulating the influence of spatial and temporal tokens.
Following the factorised attention approach of TimeSformer, this method uses different combinations of spatial and temporal attention blocks to encode the information.
These attention-based approaches have demonstrated robust performance in modelling such dependencies, facilitating more expressive and context-aware representations.
BEVPredFormer builds on this concept by applying attention mechanisms to past BEV features, thereby enhancing temporal reasoning and improving the quality of the generated predictions.

\section{Method}\label{sec3}

The overall layout of BEVPredFormer is represented in Figure~\ref{fig:architecture}.
The main components of the architecture are: (1) Image feature extraction using a backbone and neck. (2) A view transform based on self- and cross-attention. (3) A Sparse UNet block to further process BEV features. (4) A temporal block focused on exploiting temporal relations and patterns. (5) Multiple multi-scale predictor heads.

BEVPredFormer relies only on the vehicle localisation and a multi-camera system as input sources.
The location history of the vehicle is used to compensate for ego-motion throughout the input sequence, aligning all observations into a common reference frame.
The sequence of images \(I \in \mathbb{R}^{T_{in} \times N_{cam} \times 3 \times H_{im} \times W_{im}}\) is the primary source of environmental information, where \(T_{in}\) denotes the number of input frames in the sequence, \(N_{cam}\) is the number of cameras in the setup, and \(H_{im} \times W_{im}\) represents the image resolution.

The final objective is to predict an instance-level map \(P \in \mathbb{R}^{T_{out} \times H \times W}\) for \(T_{\text{out}}\) future frames.
This map is obtained from two intermediate representations: the backwards flow map \(F \in \mathbb{R}^{T_{out} \times 2 \times H \times W}\), which captures per-pixel motion in the BEV plane, and the semantic segmentation map \(S \in \mathbb{R}^{T_{out} \times N_c \times H \times W}\), where \(N_c = 2\) denotes the number of semantic classes: vehicles and background.
The BEV feature map resolution \(H \times W\) is fixed to \(200 \times 200\) cells throughout the architecture.

\begin{figure*}[t]
    \centering
    \includegraphics[width=0.7\linewidth]{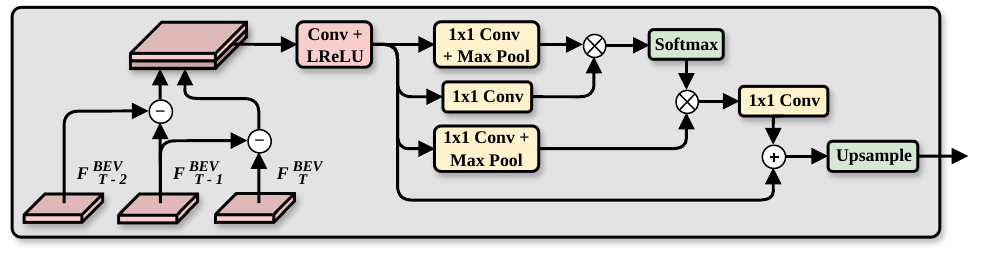}
    \caption{Difference module architecture present in the temporal block of BEVPredFormer}
    \label{fig:diff_module}
\end{figure*}

As depicted in Figure~\ref{fig:architecture}, the sequence of multi-camera images \(I\) is processed by a shared feature extractor and a neck that fuses multi-scale features.
A BEVFormer architecture is used to project past information into the BEV space.
First, the BEV queries \(Q_{BEV}\) are processed by a self-attention module, and then the final camera features are projected into the BEV plane by a cross-attention module.
A Sparse UNet block is then responsible for further refinement.
Next, the temporal module enhances the representation of temporal information by first calculating and adding the difference features, and then processing the BEV feature sequence using PredFormer blocks, to obtain the final BEV features \(F^{BEV}_{temporal}\).
Two multi-scale heads are responsible for projecting into future frames and extracting flow \(F\) and segmentation \(S\) maps, which will be needed to obtain the final instance prediction \(P\).

\subsection{Feature Extraction}\label{subsec3.1}

The first stage of BEVPredFormer focuses on extracting features from the input data. In complex architectures, reaching a balance between accuracy and computational efficiency is crucial.
SOTA models often use backbones such as EfficientNet \cite{efficientnet} or Swin Transformer \cite{liu2021swin}.
BEVPredFormer uses EfficientViT \cite{liu2023efficientvit} as its backbone, which has been specifically designed to maximise the efficiency of attention blocks through a cascaded group attention mechanism.

The input consists of a sweep of six cameras \(I\), with each image processed independently.
EfficientViT extracts multi-scale feature maps for each image with downsampling factors of 8 and 16, and channel dimensions of 128 and 256, respectively.
A neck module then combines each group of multi-scale features to create a unified feature representation for each respective image, resulting in the final extracted features, \(I_F \in \mathbb{R}^{(T_{\text{in}} \cdot N_{cam}) \times C_F \times H_F \times W_F}\).
This module includes an upsampling layer to align the spatial features, followed by two convolution–normalisation–activation blocks.

\subsection{BEV Projection and Refinement}\label{subsec3.2}

The second module of the architecture (stage 2 in Figure~\ref{fig:architecture}) is responsible for projecting all the information extracted from the input images into a unified BEV representation, which is spatially aligned with the real-world 3D environment.
By transforming multi-view image features into a top-down, flattened space, the model enhances spatial reasoning and temporal alignment across frames while reducing computational complexity.
The projection block in BEVPredFormer is based on BEVFormer, using attention mechanisms to model the relationship between images and BEV planes.

First, \(Q_{BEV} \in \mathbb{R}^{T_{in} \times C_{BEV} \times H \times W} \), a set of learnable BEV queries, is processed by a self-attention module that focuses on extracting spatial dependencies.
Second, image features are aggregated into the BEV space through a cross-attention mechanism.
In this process, each query corresponds to a point in the predefined 3D BEV grid.
Each grid position is first projected onto the image plane using the corresponding intrinsic and extrinsic parameters.
The projected points are then used to sample relevant features from the multi-view image feature maps.
The cross-attention module then facilitates interaction between these sampled features and the BEV queries, aggregating spatially aligned visual information from multiple views into the BEV representation.
Finally, a feed-forward network (FFN) processes the aggregated temporal and spatial information.
This sequence of self-attention, cross-attention and FFN is repeated \(N_{BEV}\) times, obtaining the past BEV features \(F^{BEV}\in \mathbb{R}^{T_{in} \times C_{BEV} \times H \times W}\).
Following the cross-attention projection, the BEV features are processed further using a Sparse UNet module applied to BEV features as in PointBEV \cite{chambon2024pointbev} (stage 3 in Figure~\ref{fig:architecture}).
This component improves the representation by capturing local and global spatial dependencies through downsampling and upsampling paths with skip connections.
The multi-scale aggregation enables the model to reason more effectively about object shapes and spatial context across the BEV map, obtaining a new set of BEV features \(F^{BEV}_{processed}\).

\subsection{Temporal Module}\label{subsec3.3}

BEVPredFormer incorporates a temporal module that focuses on exploiting the temporal relations between the different BEV states of the previous and current frames.
This stage is divided into two fundamental blocks: a difference module and a PredFormer encoder.

\subsubsection{Difference Module}\label{subsubsec3.3.1}

BEVPredFormer incorporates a difference module similar to the one presented in DMP.
The architecture, represented in Figure~\ref{fig:diff_module}, begins by computing difference features as the frame-wise subtraction of consecutive BEV feature maps from the input sequence $T$, $T-1$ and $T-2$.
These difference features capture temporal variations in the scene, which are often caused by moving objects.
To highlight regions with significant changes, an additional weighting mechanism assigns greater value to areas with large difference features. These weights are combined with the original difference features and then upsampled to obtain \(F^{BEV}_{diff}\), the corresponding past BEV features with enhanced flow representation.

\subsubsection{PredFormer Blocks}\label{subsubsec3.3.2}

\begin{figure}[t]
    \centering
    \includegraphics[width=0.95\linewidth]{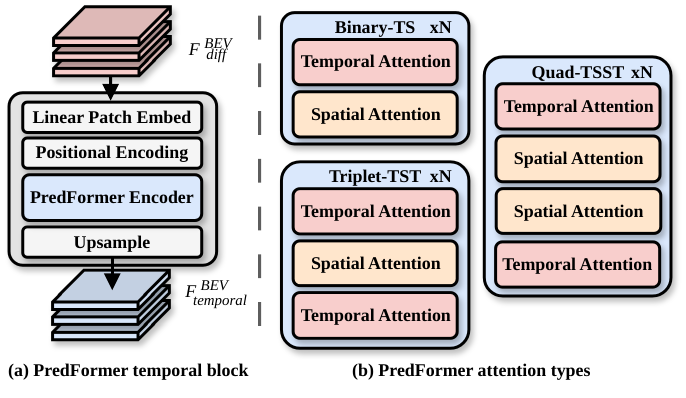}
    \caption{Temporal attention architecture. (\textbf{a}) Overall diagram of spatio-temporal processing. (\textbf{b}) PredFormer attention types}
    \label{fig:predformer_module}
\end{figure}

After the motion-guided refinement stage, the proposed architecture uses the implementation of PredFormer blocks represented in Figure~\ref{fig:predformer_module}a to process the sequence of past BEV difference-guided feature maps.
PredFormer, which was originally designed for video prediction, specialises in capturing spatio-temporal dependencies by combining spatial and temporal self-attention within gated transformer layers.
In our adaptation, these blocks use the BEV difference feature maps \(F^{BEV}_{diff}\) as input, enabling the modelling of temporal consistency and cross-frame context directly in BEV space.
This enables the network to learn dynamic patterns and improve prediction performance over time.

First, the BEV feature map is divided into non-overlapping patches by a patch embedding layer and projected into a lower-dimensional token sequence, reducing computational overhead.
Next, a set of prediction queries \(Q_{pred}\) is generated for the attention mechanism by applying linear projections to \(F^{BEV}_{diff}\).
Absolute sinusoidal positional encodings are then added to these tokens to preserve spatial coherence across the map.
The transformer-based PredFormer block is applied multiple times in sequence to capture complex spatio-temporal dependencies.
Similarly to the original PredFormer, in order to improve the flow of information inside the temporal module, the attention uses a GTB, replacing the Multi-Layer Perceptron (MLP) with a Gated Linear Unit (GLU).
Finally, an upsampling layer is added to reproject the processed tokens back into the full spatial map scale to restore the original BEV resolution and generate the final feature map \(F^{BEV}_{temporal} \in \mathbb{R}^{T_{in} \times C_{BEV} \times H \times W}\).

As shown in Figure~\ref{fig:predformer_module}b, PredFormer blocks support multiple attention configurations, breaking down spatio-temporal processing into distinct spatial and temporal stages. This divided attention strategy improves learning efficiency and performance by separating the spatial structure from the temporal dynamics. In our experiments, applying temporal attention first yielded slightly higher accuracy, although this difference is small and is considered an exploratory observation (Section~\ref{subsec4.4}). For this reason, our architecture focuses on three specific configurations: Binary-TS, Triplet-TST and Quadruplet-TSST, where T and S denote a temporal and a spatial attention layer, respectively, listed in the order in which they are applied.

\subsection{Flow and Segmentation Heads}\label{subsec3.4}

The temporally enhanced BEV features \(F^{BEV}_{temporal}\) are then passed to the last stage of the architecture to generate the final prediction.
In line with the approach proposed in \cite{powerbev}, our method produces two output representations: a backwards optical flow map \(F \in \mathbb{R}^{T_{out} \times 2 \times H \times W}\) and a semantic segmentation map \(S \in \mathbb{R}^{T_{out} \times N_c \times H \times W}\).

The backwards flow map \(F\) represents the displacement of each pixel towards the centre of the same instance in the previous frame.
This formulation mitigates errors in propagation that are more likely to occur when using forward flow, particularly when there are fast-moving objects or occlusions present.
The segmentation map \(S\) captures the occupancy of the BEV grid, indicating which cells are occupied by vehicles in each predicted future frame.

A post-processing step is applied to obtain the final instance prediction map, \(P = P(S, F)\). For each predicted frame, the segmentation map \(S\) is used to identify the BEV positions occupied by vehicles, acting as a spatial mask. At the same time, the backwards flow map \(F\) is used to propagate instance IDs from the last predicted frame through the output sequence.

In terms of head structure, we adopt a multi-scale approach \cite{powerbev,dmp}, composed of an encoder that uses \(F^{BEV}_{temporal}\) to generate features at different scales, a predictor module that projects each scale into future space \(T_{out}\), and a decoder that fuses all future information for the final prediction.
As explained in Section~\ref{subsec4.1}, the number of output maps \(T_{out}\) depends on the number of seconds into the future that are being predicted.
During training, we use two lightweight auxiliary heads to incorporate additional supervision, predicting centerness and offset maps.

\section{Experiments}\label{sec4}

\subsection{Implementation and Datasets}\label{subsec4.1}

To perform a fair comparison with other SOTA methods, the experiments are conducted on the nuScenes dataset \cite{nuscenes2019}, following the official split of 1000 driving scenes: 700 for training, 150 for validation and 150 for testing.
All performance metrics are reported on the validation set, using the original sensor data at a frequency of 2 Hz, following the standard protocol of previous Instance Prediction works \cite{fiery2021,powerbev,beverse}. The test set cannot be used for this task, since its annotations are not publicly available and the official nuScenes benchmark does not include Instance Prediction.
Following the evaluation configuration of other SOTA methods, we restrict the dataset labels to objects categorised as vehicles and with a minimum camera visibility of 40\%.
In terms of the temporal configuration, we follow the standard convention of using input information from the present moment and one second in the past to make predictions two seconds into the future.
With a frame rate of 2 Hz, this corresponds to three input frames \(T_{in} = 3\) and four output frames \(T_{pred} = 4\), needing two additional future frames to initialise and propagate instance representations \(T_{out} = T_{pred} + 2 = 6\).
We adopt a fixed BEV resolution of \(H \times W = 200 \times 200\) and \(C_{BEV} = 128\) channels.
The experiments consider two spatial ranges: a short range of 15 m and a long range of 50 m from the ego vehicle.
The temporal module uses a patch size of 4 to process the BEV features.
Following PowerBEV \cite{powerbev}, instances are obtained by thresholding the predicted centres at 0.1, applying non-maximum suppression with a kernel size of 5 and 13 cells for the long and short ranges, respectively, assigning each foreground pixel to its nearest centre, and propagating the identities to future frames by warping them with the predicted backwards flow. The vehicle class includes all nuScenes vehicle subcategories.

\subsection{Metrics and Losses}\label{subsec4.2}

The different experiments follow the SOTA evaluation methodology and report two primary metrics: Intersection over Union (IoU) and Video Panoptic Quality (VPQ).

The IoU metric, defined in Equation~(\ref{eq:iou}), is a standard metric used to evaluate segmentation performance.
In the context of BEV prediction, the IoU measures the overlap between the predicted and ground-truth vehicle occupancy maps for each subsequent frame.
The overall IoU metric is computed over both the present and future \(T_\text{pred}\) frames, accumulating the intersection and union of all frames and all sequences of the validation set before computing the ratio, where \(|\cdot|\) denotes the number of BEV cells.
This makes it possible to evaluate the accuracy with which the model identifies the spatial extent of objects in the scene.
However, IoU is limited to per-frame evaluations and does not consider the temporal consistency of instance predictions.
\begin{equation}
    \text{IoU} = \frac{\sum_{t=0}^{T_\text{pred}} |\hat{y}_{t}^\text{seg} \cap {y}_{t}^\text{seg}|}{\sum_{t=0}^{T_\text{pred}} |\hat{y}_{t}^\text{seg} \cup {y}_{t}^\text{seg}|}
    \label{eq:iou}
\end{equation}

To address this limitation, this study employs VPQ, defined in Equation~(\ref{eq:vpq}), which extends the traditional Panoptic Quality (PQ) metric to video. VPQ evaluates the quality of instance predictions over time by considering detection, segmentation (Segmentation Quality) and identity consistency (Recognition Quality).
VPQ penalises fragmented tracks or inconsistent segmentations of the same object across frames, making it particularly well-suited to evaluating temporal models.
It is computed as
\begin{equation}
    \text{VPQ} = \frac{\sum_{t=0}^{T_\text{pred}} \sum_{(p_t,q_t) \in TP_t} \text{IoU}(p_t,q_t)}{\sum_{t=0}^{T_\text{pred}} \left( |TP_t| + \frac{1}{2}|FP_t| + \frac{1}{2}|FN_t| \right)}
    \label{eq:vpq}
\end{equation}
where \(TP_t\), \(FP_t\) and \(FN_t\) denote the sets of true positive, false positive and false negative instances at frame \(t\), respectively.
A predicted instance \(p_t\) is matched with a ground-truth instance \(q_t\) when their IoU is greater than 0.5, which guarantees a unique matching.
To evaluate identity consistency, each ground-truth instance keeps track of the ID of the prediction it was matched with in previous frames of the sequence; if it is matched with a prediction with a different ID, the pair is counted as a false positive and a false negative instead of a true positive.
All terms are accumulated over every frame and every sequence of the validation set before computing the ratio, and only the vehicle class is considered.

In addition to flow and segmentation, two auxiliary tasks are also supervised during training to improve the prediction performance.
The segmentation loss \(\mathcal{L}_\text{seg}\) uses a cross-entropy loss computed over the 25\% of pixels with the highest loss, which focuses training on hard examples and reduces the dominance of the background class, with class weights of 1 and 2 for the background and vehicle classes, respectively, and a temporal discount of \(0.95^t\) over the output frames.
Flow is trained using a smooth L1 loss \(\mathcal{L}_\text{flow}\) (\(\beta = 1\)) to provide a balance of robustness and precision in motion estimation.
The centerness loss \(\mathcal{L}_\text{center}\) is optimised using a mean squared error (MSE) loss, while per-pixel offsets \(\mathcal{L}_\text{off}\) are trained using an L1 loss.
All loss components are combined in Equation~(\ref{eq:loss}) using dynamic weighting, which allows the model to adaptively balance the contribution of each task during training. Following the uncertainty-based weighting of \cite{kendall2018multi}, each loss term \(\mathcal{L}_i\) is scaled by \(e^{-s_i}\) and regularised by \(s_i\), where \(s_i\) is a learnable log-variance initialised to zero and optimised jointly with the network, and \(N = 4\) is the number of loss terms.

\begin{align}
    \mathcal{L} &= \frac{1}{N} \sum_{i=1}^{N} \left( \frac{1}{2} e^{-s_i} \mathcal{L}_i + \frac{1}{2} s_i \right), \notag \\
    \mathcal{L}_i &\in \{\mathcal{L}_\text{seg}, \mathcal{L}_\text{flow}, \mathcal{L}_\text{off}, \mathcal{L}_\text{center}\}
    \label{eq:loss}
\end{align}

\subsection{Results}\label{subsec4.3}

\begin{table*}[t]
\caption{Comparative model performance on nuScenes validation set. All metrics are obtained using the official model implementation, if available. $^\dagger$ No official implementation available; results as reported in the original paper. Unless otherwise indicated, all methods use an input resolution of \(224 \times 480\). $^\star$ Input resolution of \(448 \times 800\). $^\diamond$ Input resolution of \(256 \times 704\). Best value is marked in \textbf{bold}.}
\label{tab:comparative-sota}
\begin{tabular*}{\textwidth}{@{\extracolsep\fill}lcccccc@{}}
\toprule
\multirow{2}{*}{\textbf{Architecture}} & \multirow{2}{*}{\textbf{Encoder}} & \multirow{2}{*}{\textbf{View Transform}} & \multicolumn{2}{c}{\textbf{Long Range}} & \multicolumn{2}{c}{\textbf{Short Range}} \\ \cmidrule(lr){4-5} \cmidrule(lr){6-7}
 & & & \textbf{VPQ $\uparrow$} & \textbf{IoU $\uparrow$} & \textbf{VPQ $\uparrow$} & \textbf{IoU $\uparrow$} \\ \midrule
PowerBEV \cite{powerbev}           & EfficientNet & Lift-Splat         & 32.2 & 38.9 & 55.5 & 62.5 \\
Fiery \cite{fiery2021}             & EfficientNet & Lift-Splat         & 29.9 & 36.7 & 50.2 & 59.4 \\
StretchBEV \cite{stretchbev}       & EfficientNet & Lift-Splat         & 29.0 & 37.1 & 46.0 & 55.5 \\
BEVerse$^\diamond$ \cite{beverse}  & Swin         & Lift-Splat         & 33.3 & 38.7 & 52.2 & 60.3 \\
ST-P3 \cite{stp3}                  & EfficientNet & Lift-Splat         & 32.0 & 38.9 & -    & -    \\
FaE \cite{fast_effic}              & EfficientNet & Lift-Splat         & 29.8 & 37.4 & 53.7 & 59.1 \\
DMP$^\dagger$ \cite{dmp}           & Swin         & Lift-Splat + Attn. & 34.0 & 38.8 & \textbf{57.5} & 62.9 \\ \midrule
BEVPredFormer                      & EfficientViT & BEVFormer          & 34.1 & 41.0 & 53.8 & 62.7 \\
BEVPredFormer$^\star$              & EfficientViT & BEVFormer          & \textbf{35.4} & \textbf{43.3} & 54.9 & \textbf{64.0} \\ \bottomrule
\end{tabular*}
\end{table*}

The architecture is trained using a two-stage process. The first stage involves adapting the image feature extractor and BEV projection components to the nuScenes dataset.
For this purpose, the model is trained for 50 epochs to perform BEV segmentation on the current frame without making any predictions.
The input is the same past sequence data used for the full task, maintaining compatibility with the temporal setup of the second stage.
In the second stage, the focus shifts to predicting future instances.
The pre-trained weights from the first stage, corresponding to the image feature extractor, BEV projector and Sparse UNet, are loaded and kept frozen.
The remaining modules are then trained for an additional 50 epochs.
This staged approach results in faster convergence during the temporal training phase, improving final prediction accuracy.
To assess the impact of this strategy, BEVPredFormer was also trained end-to-end without the segmentation pre-training stage, using the same total budget of 100 epochs and obtaining a long-range VPQ of 35.2\% and IoU of 42.8\%. Compared with the two-stage training, VPQ and IoU decrease by 0.2 and 0.5 points, respectively. The IoU decrease clearly exceeds the seed-to-seed variability measured in Section~\ref{subsec4.4}, whereas the VPQ decrease is within the range regarded as exploratory. Nevertheless, the end-to-end model still outperforms all compared methods in the long-range setting.
The compared methods in Table~\ref{tab:comparative-sota} are evaluated using their official implementations. When retraining Fiery, PowerBEV and BEVerse for up to 50 epochs, we observed that their performance saturates at around 25--30 epochs, so they are evaluated at convergence. The longer schedule of BEVPredFormer is required by its attention-based design, which converges more slowly than convolutional pipelines.

\begin{table*}[b!]
\caption{Computational cost of the compared methods, measured with re-implementations within a common framework on a single NVIDIA RTX 5090 GPU (batch size 1, FP32). Details of the measurement protocol are given in Section~\ref{subsec4.3}. $^\ddagger$ Motion prediction branch of BEVerse-Tiny.}
\label{tab:efficiency}
\begin{tabular*}{\textwidth}{@{\extracolsep\fill}lcccccc@{}}
\toprule
\textbf{Architecture} & \tabhead{\textbf{Input}\\\textbf{Resolution}} & \tabhead{\textbf{Params}\\\textbf{(M)}} & \textbf{GFLOPs} & \tabhead{\textbf{Peak}\\\textbf{Memory (GB)}} & \tabhead{\textbf{Forward}\\\textbf{(ms) $\downarrow$}} & \tabhead{\textbf{End-to-End}\\\textbf{(ms) $\downarrow$}} \\ \midrule
PowerBEV \cite{powerbev}            & \(224 \times 480\) & 39.1  & 203  & 1.07 & 24.6 & 59.4 \\
Fiery \cite{fiery2021}              & \(224 \times 480\) & 8.3   & 501  & 0.92 & 34.5 & 69.8 \\
StretchBEV \cite{stretchbev}        & \(224 \times 480\) & 17.0  & 663  & 0.96 & 32.4 & 65.2 \\
BEVerse$^\ddagger$ \cite{beverse}   & \(256 \times 704\) & 62.5  & 4139 & 1.42 & 74.0 & 124.4 \\
FaE \cite{fast_effic}               & \(224 \times 480\) & 13.4  & 142  & 0.94 & 23.5 & 57.8 \\ \midrule
BEVPredFormer                       & \(224 \times 480\) & 109.5 & 2004 & 1.66 & 57.7 & 95.9 \\
BEVPredFormer                       & \(448 \times 800\) & 109.5 & 2915 & 2.30 & 83.2 & 167.6 \\ \bottomrule
\end{tabular*}
\end{table*}

The BEVPredFormer model is trained on two NVIDIA A100 GPUs with 80 GB of VRAM.
The experiments use the AdamW optimiser with a maximum learning rate of \(3 \times 10^{-4}\), scheduled using the One Cycle learning rate policy with an initial learning rate of \(10^{-4}\) and a 5\% warm-up, an effective batch size of 16, gradient clipping at a norm of 5 and 32-bit floating-point (FP32) precision.
The backbone is initialised with ImageNet-1k pre-trained weights.

Unless otherwise specified, all input images are resized to a resolution of \(448 \times 800\) as part of the pre-processing pipeline.
To improve the robustness and generalisation of the model, we apply several data augmentation techniques:
(1) Random zoom and rotation are applied to the input multi-camera images, with factors in \([0.95, 1.05]\) and angles in \([-2.5^\circ, 2.5^\circ]\).
(2) Random translations and rotations are applied in the BEV representation coherently across the entire input sequence in order to preserve spatial and temporal consistency, with translations of up to \(\pm 30\) m and \(\pm 20\) m along the \(x\) and \(y\) axes and rotations of up to \(\pm 20^\circ\) around the vertical axis.

% Short range results in scenes 200, 4464, 4036, 1072
\begin{figure*}[h]
    \centering
    \includegraphics[width=0.80\linewidth]{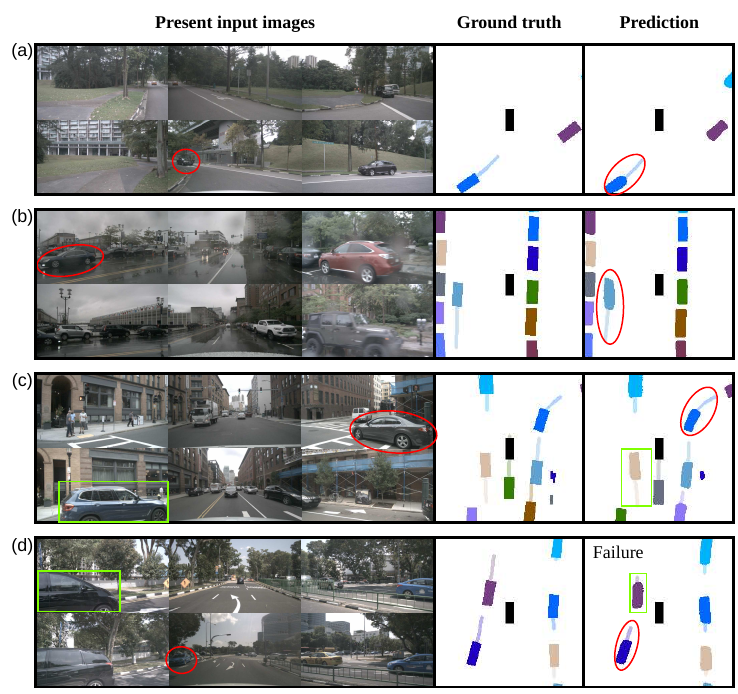}
    \caption{BEVPredFormer short-range results. Different highlighted areas represent the same object in the camera and BEV. The different scenes include: (\textbf{a}) Curved road prediction; (\textbf{b}) Rainy weather; (\textbf{c}) Intersection; (\textbf{d}) Prediction failure.}
    \label{fig:qualitative_short}
\end{figure*}

Table~\ref{tab:comparative-sota} compares BEVPredFormer with Instance Prediction SOTA models, all of which are benchmarked on the nuScenes dataset. The evaluated methods use different backbones, with most relying on EfficientNet variants for image encoding, whereas BEVPredFormer uses EfficientViT.
Most SOTA approaches project image features into 3D space using BEV transformations based on LSS, which rely on depth estimation and volumetric projection.
By contrast, BEVPredFormer builds on BEVFormer to perform a direct 2D-to-BEV transformation, eliminating the need for explicit depth estimation. As shown in Section~\ref{subsec4.4}, this attention-based projection improves accuracy and reduces latency compared with LSS.
Using an input resolution of \(448 \times 800\), BEVPredFormer achieves SOTA results in three of the four key metrics. For long-range Instance Prediction, it obtains a VPQ score of 35.4\%, outperforming all compared methods, including DMP (34.0\%) and BEVerse (33.3\%).
In terms of segmentation accuracy, BEVPredFormer achieves a long-range IoU of 43.3\%, surpassing the next best methods, PowerBEV and ST-P3, by 4.4 points. In the short-range setting, it reaches the highest IoU score of 64.0\% and a VPQ score of 54.9\%, which is competitive with PowerBEV (55.5\%) and surpassed only by DMP (57.5\%), which lacks an official open-source implementation.
When using the same \(224 \times 480\) input resolution as the compared methods, BEVPredFormer still obtains the best long-range IoU of 41.0\%, surpassing the next best methods by 2.1 points, and a long-range VPQ of 34.1\%, comparable to DMP (34.0\%) and higher than all methods with an official implementation. In the short-range setting, it achieves an IoU of 62.7\%, only below DMP (62.9\%), and a VPQ of 53.8\%, below PowerBEV and DMP. In terms of computational cost (Table~\ref{tab:efficiency}), BEVPredFormer requires 57.7 ms per forward pass at this resolution, which is lower than BEVerse (74.0 ms) and with fewer giga floating-point operations (GFLOPs), but higher than lightweight LSS-based models such as FaE (23.5 ms), PowerBEV (24.6 ms), StretchBEV (32.4 ms) and Fiery (34.5 ms), which also have fewer parameters. This additional cost is reflected in a consistent improvement in long-range performance, where BEVPredFormer surpasses all these models in both VPQ and IoU. Increasing the resolution to \(448 \times 800\) raises the latency to 83.2 ms, but improves all four accuracy metrics, especially in the long-range setting.
Table~\ref{tab:efficiency} also reports the peak GPU memory and the end-to-end latency of each method.
To ensure a fair comparison, all computational metrics are measured using reimplementations of each method within a common framework. The same preprocessing and postprocessing are used for all methods, and the results are averaged over 50 runs after a warm-up phase.
Where available, the parameter counts of these re-implementations match those of the official checkpoints. The values reported for BEVerse correspond to the motion prediction branch of BEVerse-Tiny, excluding its 3D detection and map segmentation branches.
Forward latency only includes the model forward pass. End-to-end latency, on the other hand, is measured as a separate sequential pipeline that includes image preprocessing of decoded images in host memory. This includes resizing, cropping, normalization, camera and ego-motion parameters, and host-to-device transfer. It also includes the post-processing required to obtain the final instance prediction.
Since post-processing time depends on the number of predicted instances, it is measured using the same reference prediction for all methods. Peak memory refers to the maximum amount of GPU memory allocated during inference.
All methods require less than 2.5 GB of GPU memory, with BEVPredFormer requiring 1.7 and 2.3 GB at \(224 \times 480\) and \(448 \times 800\), respectively.
Post-processing takes less than 2 ms for all methods, whereas preprocessing, which is performed on the CPU and grows with the input resolution, accounts for a large part of the end-to-end latency, particularly for the lightweight models, and could be accelerated by performing it on the GPU. At \(224 \times 480\), the end-to-end latency of BEVPredFormer is 95.9 ms, 36.5 ms higher than PowerBEV, a difference similar to that of the forward pass.
At \(448 \times 800\), it increases to 167.6 ms, which still corresponds to about 6 predictions per second, above the 2 Hz frequency of the nuScenes keyframes used as input.

% Long range results in scenes 140, 1976, 4808, 5108
\begin{figure*}[t]
    \centering
    \includegraphics[width=0.80\linewidth]{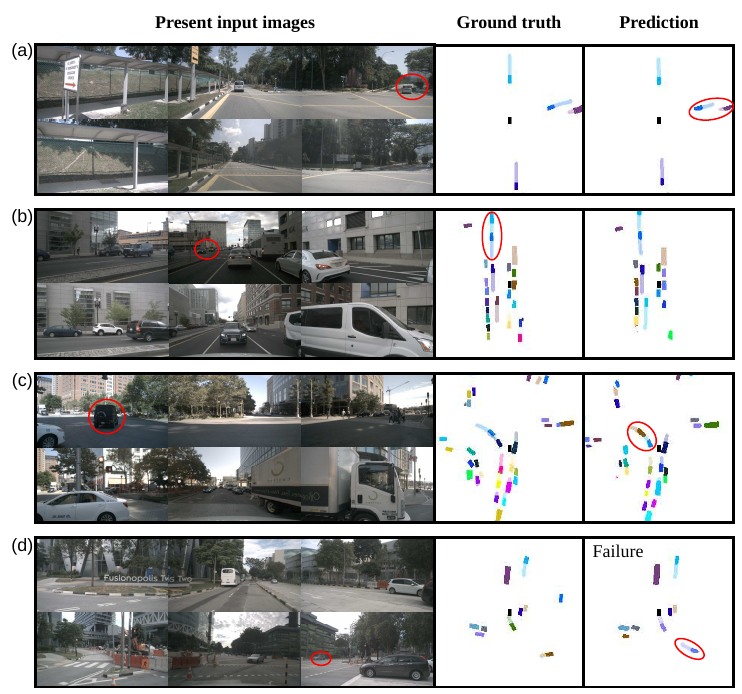}
    \caption{BEVPredFormer long-range results. Different highlighted areas represent the same object in the camera and BEV. The different scenes include: (\textbf{a}) Long-range prediction; (\textbf{b}) Crowded street; (\textbf{c}) Intersection; (\textbf{d}) Prediction failure.}
    \label{fig:qualitative_long}
\end{figure*}

Figure~\ref{fig:qualitative_short} presents qualitative results of BEVPredFormer on a set of nuScenes validation scenarios using the short-range configuration.
The black box at the centre of each scene represents the ego-vehicle position at the current frame.
In (a), the model correctly predicts a moving vehicle following the ego vehicle along a curved road.
In (b), BEVPredFormer accurately segments all parked vehicles on both sides of the road, as well as a moving vehicle approaching from the opposite direction under rainy conditions.
The more complex urban scenario in (c) shows that the model correctly identifies moving vehicles in both directions and predicts a right turn at the intersection.
However, in (d), although all vehicle instances are correctly segmented, the model fails to accurately predict their long-term motion, underestimating the displacement of the highlighted vehicles. In particular, the vehicle marked in green is split between the front-left and back-left cameras, and these partial observations from different views likely hinder the estimation of its velocity.

\begin{table*}[h]
\caption{Performance of BEVPredFormer (\(448 \times 800\)) by ground-truth vehicle speed \(v\) at the present frame. Unmatched predictions are assigned to the group of the ground-truth vehicle they overlap or, otherwise, according to their predicted speed.}
\label{tab:speed}
\begin{tabular*}{\textwidth}{@{\extracolsep\fill}lcccccc@{}}
\toprule
 & \multicolumn{3}{c}{\textbf{Long Range}} & \multicolumn{3}{c}{\textbf{Short Range}} \\ \cmidrule(lr){2-4} \cmidrule(lr){5-7}
\textbf{Speed (m/s)} & \textbf{VPQ $\uparrow$} & \textbf{SQ $\uparrow$} & \textbf{Recall $\uparrow$} & \textbf{VPQ $\uparrow$} & \textbf{SQ $\uparrow$} & \textbf{Recall $\uparrow$} \\ \midrule
Static (\(v < 0.5\))         & 41.8 & 72.6 & 54.9 & 63.6 & 77.4 & 81.8 \\
Slow (\(0.5 \leq v \leq 5\)) & 19.2 & 69.1 & 42.7 & 36.9 & 73.0 & 60.0 \\
Fast (\(v > 5\))             & 30.1 & 68.9 & 41.0 & 30.9 & 71.4 & 34.8 \\ \bottomrule
\end{tabular*}
\end{table*}

\begin{table*}[h]
\caption{Mean and standard deviation over three matched random seeds of the model without temporal module and three temporal configurations (two blocks, patch size 4), and difference of each temporal configuration with respect to the model without temporal module, with its 95\% CI in brackets. $^\dagger$ Temporal module without difference block.}
\label{tab:seeds}
\setlength{\tabcolsep}{3pt}%
\begin{tabular*}{\textwidth}{@{\extracolsep\fill}lcccc@{}}
\toprule
\multirow{2}{*}{\textbf{Configuration}} & \multicolumn{2}{c}{\textbf{Long Range}} & \multicolumn{2}{c}{\textbf{Short Range}} \\ \cmidrule(lr){2-3} \cmidrule(lr){4-5}
 & \textbf{VPQ $\uparrow$} & \textbf{IoU $\uparrow$} & \textbf{VPQ $\uparrow$} & \textbf{IoU $\uparrow$} \\ \midrule
No temporal            & \(34.58 \pm 0.03\) & \(42.86 \pm 0.04\) & \(54.77 \pm 0.02\) & \(63.36 \pm 0.02\) \\
Binary-TS              & \(34.94 \pm 0.04\) & \(43.02 \pm 0.07\) & \(54.89 \pm 0.01\) & \(63.83 \pm 0.04\) \\
Triplet-TST$^\dagger$  & \(35.06 \pm 0.03\) & \(42.90 \pm 0.05\) & \(54.39 \pm 0.02\) & \(63.81 \pm 0.03\) \\
Triplet-TST            & \(35.32 \pm 0.05\) & \(43.21 \pm 0.07\) & \(54.84 \pm 0.06\) & \(63.87 \pm 0.05\) \\ \midrule
\multicolumn{5}{@{}l}{\textit{Difference with respect to No temporal [95\% CI]}} \\
Binary-TS              & \(+0.36\) \([+0.24, +0.48]\) & \(+0.16\) \([-0.04, +0.36]\) & \(+0.12\) \([+0.06, +0.18]\) & \(+0.47\) \([+0.36, +0.58]\) \\
Triplet-TST$^\dagger$  & \(+0.48\) \([+0.37, +0.59]\) & \(+0.04\) \([-0.12, +0.20]\) & \(-0.38\) \([-0.45, -0.31]\) & \(+0.45\) \([+0.36, +0.54]\) \\
Triplet-TST            & \(+0.74\) \([+0.60, +0.88]\) & \(+0.35\) \([+0.15, +0.55]\) & \(+0.07\) \([-0.09, +0.23]\) & \(+0.51\) \([+0.38, +0.64]\) \\ \bottomrule
\end{tabular*}
\end{table*}

The results for the long-range configuration are presented in Figure~\ref{fig:qualitative_long}.
In (a), the model correctly predicts vehicles at medium and long ranges throughout the intersection.
The urban scenarios in (b) and (c) further demonstrate accurate segmentation and motion prediction for parked, moving, and turning vehicles at different distances from the ego vehicle.
Finally, (d) shows that BEVPredFormer can still struggle with long-term motion prediction in some cases, particularly for vehicles located farther from the ego vehicle. In this scene, the highlighted vehicle covers only a few pixels in the back-right camera, and this low visibility, combined with the complex layout of the intersection, leads to an inaccurate prediction of its trajectory.

To further analyse these failure cases, Table~\ref{tab:speed} reports the performance of BEVPredFormer for vehicles grouped by their ground-truth speed at the present frame. Besides VPQ, we report the segmentation quality (SQ), i.e., the mean IoU of matched instances, and the recall, i.e., the fraction of ground-truth instances that are correctly matched. Static vehicles, which account for most instances, obtain the best results in all metrics. Moving vehicles are segmented with a similar SQ once matched, but they are harder to detect and track consistently, with a lower recall, since errors in the predicted motion accumulate over the prediction horizon. This is most noticeable for fast vehicles in the short-range setting, whose finer grid makes small displacement errors more critical. In the long-range setting, slow vehicles obtain the lowest VPQ, mainly due to a large number of false positives that do not overlap any ground-truth vehicle and are predicted with a low speed. This suggests that the model struggles in ambiguous situations where vehicles may start or stop moving, as its deterministic output cannot represent several possible futures.

\subsection{Ablation Studies}\label{subsec4.4}

To evaluate the accuracy of the main comparisons in this section, four configurations were trained using the same three random seeds in the long- and short-range settings: the model without a temporal module, the temporal module with Binary-TS blocks, the temporal module with Triplet-TST blocks but without the difference module, and the default configuration (two Triplet-TST blocks with the difference module).
Table~\ref{tab:seeds} reports the mean and standard deviation of each configuration, together with the difference between each temporal configuration and the model without a temporal module and its 95\% confidence interval (CI), computed from the standard deviations of both configurations using a Student's t distribution with two degrees of freedom.
All three temporal configurations improve the long-range VPQ and the short-range IoU with CIs that exclude zero, but only the default configuration also achieves a significant gain in long-range IoU, of 0.35 points.
In contrast, the short-range VPQ improves by only 0.12 points with Binary-TS and 0.07 points with the default configuration, the latter with a CI that includes zero. This indicates that the temporal module primarily improves vehicle segmentation rather than temporal consistency at short range.

The same procedure is used to compare the temporal configurations directly.
Adding the difference module to Triplet-TST improves the long-range VPQ and IoU by 0.26 (95\% CI \([+0.12, +0.40]\)) and 0.31 points (\([+0.10, +0.52]\)), respectively, and the short-range VPQ by 0.45 points (\([+0.29, +0.61]\)), while the short-range IoU difference of 0.06 points (\([-0.08, +0.20]\)) is not significant.
Without the difference module, the short-range VPQ falls 0.38 points below that of the model without a temporal module, which suggests that the difference features help the temporal attention to preserve the temporal consistency of nearby instances.
Compared with Binary-TS, Triplet-TST improves the long-range VPQ by 0.38 points (\([+0.22, +0.54]\)), whereas the differences in long-range IoU and in both short-range metrics are not significant. Therefore, the additional temporal attention layer mainly benefits the temporal consistency of long-range predictions.
The remaining ablations in this section are based on a single training run. Since the seed-to-seed standard deviation does not exceed 0.07 points in any configuration or metric, differences of 0.3 points or more clearly exceed this variability, whereas differences of 0.1--0.2 points are regarded as exploratory observations.

Table~\ref{tab:pred_type} summarises the impact of different temporal attention configurations on overall model performance. The patch size is fixed at 4, corresponding to an effective BEV resolution of 50 cells, while two temporal attention blocks are used.
Using temporal attention improves the architecture's predictive capabilities, particularly in the long-range setting, where every configuration outperforms the baseline without temporal attention.
For example, using Binary-TS attention results in gains of 0.4 points in VPQ and 0.2 points in IoU, which is consistent with the repeated runs (Table~\ref{tab:seeds}).
More sophisticated temporal attention mechanisms, such as Triplet-TST, can further enhance performance, with improvements of up to 0.8 points in VPQ and 0.4 points in IoU.
Regarding the order of the attention layers, the temporal-first variants obtain slightly better results than their spatial-first counterparts. Binary-TS and Triplet-TST improve the long-range VPQ by 0.2 points and the IoU by 0.1 points over Binary-ST and Triplet-STS, respectively, while obtaining the same short-range VPQ and a slightly higher IoU. However, since these differences fall within the range regarded as exploratory and are much smaller than the gain provided by the temporal module, they are regarded as an exploratory observation rather than evidence that the temporal-first order is superior. The temporal-first order is therefore adopted in the PredFormer blocks (Section~\ref{subsubsec3.3.2}), although a spatial-first order is expected to perform similarly.
\begin{table}[t]
\caption{Temporal module block type comparative performance. All experiments with patch size 4 and 2 blocks. $^\dagger$ Temporal module without difference block. Best value is marked in \textbf{bold}.}
\label{tab:pred_type}
\setlength{\tabcolsep}{3pt}%
\begin{tabular*}{\columnwidth}{@{\extracolsep\fill}lcccc@{}}
\toprule
\multirow{2}{*}{\textbf{Block Type}} & \multicolumn{2}{c}{\textbf{Long Range}} & \multicolumn{2}{c}{\textbf{Short Range}} \\ \cmidrule(lr){2-3} \cmidrule(lr){4-5}
 & \textbf{VPQ $\uparrow$} & \textbf{IoU $\uparrow$} & \textbf{VPQ $\uparrow$} & \textbf{IoU $\uparrow$} \\ \midrule
No temporal            & 34.6          & 42.9          & 54.8          & 63.4 \\
Binary-ST              & 34.8          & 43.0          & 54.9          & 63.7 \\
Binary-TS              & 35.0          & 43.1          & 54.9          & 63.8 \\
Triplet-STS            & 35.2          & 43.2          & 54.9          & 63.8 \\
Triplet-TST$^\dagger$  & 35.1          & 43.0          & 54.4          & 63.9 \\
Triplet-TST            & \textbf{35.4} & \textbf{43.3} & 54.9          & 64.0 \\
Quadruplet-TSST        & \textbf{35.4} & 43.2          & \textbf{55.0} & \textbf{64.1} \\ \bottomrule
\end{tabular*}
\end{table}

\begin{table}[t]
\caption{Long-range performance comparison of the number of temporal processing blocks. All experiments use the Triplet-TST block with patch size 4. GFLOPs and parameters correspond only to the temporal module. Best value is marked in \textbf{bold}.}
\label{tab:n_temp_blocks}
\setlength{\tabcolsep}{3pt}%
\begin{tabular*}{\columnwidth}{@{\extracolsep\fill}ccccc@{}}
\toprule
\textbf{N Block} & \textbf{VPQ $\uparrow$} & \textbf{IoU $\uparrow$} & \textbf{GFLOPs} & \tabhead{\textbf{Params}\\\textbf{(M)}} \\ \midrule
1 & 34.8          & 42.9          & 248 & 6.7  \\
2 & \textbf{35.4} & 43.3          & 295 & 9.9  \\
4 & 35.2          & 43.3          & 390 & 16.2 \\
6 & 35.3          & \textbf{43.5} & 485 & 22.5 \\ \bottomrule
\end{tabular*}
\end{table}

\begin{table*}[!b]
\caption{Long-range performance for different combinations of image feature extractor and view transform. All variants use the same temporal module (two Triplet-TST blocks with the difference module), prediction heads, input resolution (\(448 \times 800\)) and training schedule. LSS uses depth bins from 2 to 50~m with a 1~m resolution. Parameters, GFLOPs and forward latency are measured as in Table~\ref{tab:efficiency}. Best value is marked in \textbf{bold}.}
\label{tab:backbone_vt}
\begin{tabular*}{\textwidth}{@{\extracolsep\fill}llccccc@{}}
\toprule
\textbf{Feature Extractor} & \textbf{View Transform} & \textbf{VPQ $\uparrow$} & \textbf{IoU $\uparrow$} & \tabhead{\textbf{Params}\\\textbf{(M)}} & \textbf{GFLOPs} & \tabhead{\textbf{Latency}\\\textbf{(ms) $\downarrow$}} \\ \midrule
EfficientNet-B4 & LSS       & 30.0          & 39.2          & 99.8  & 1753 & 120.0 \\
EfficientNet-B4 & BEVFormer & 32.7          & 40.8          & 105.3 & 1936 & 109.5 \\
EfficientViT    & LSS       & 31.8          & 39.6          & 104.1 & 2732 & 87.0 \\
EfficientViT    & BEVFormer & \textbf{35.4} & \textbf{43.3} & 109.5 & 2915 & \textbf{83.2} \\ \bottomrule
\end{tabular*}
\end{table*}
In the short-range scenario, the VPQ differences between configurations are small, ranging from a 0.4-point drop when Triplet-TST is used without the difference block to a 0.2-point increase with Quadruplet-TSST. The repeated runs (Table~\ref{tab:seeds}) confirm both the small short-range VPQ gain of the configurations with the difference module and the drop without it.
In contrast, the short-range IoU improves by up to 0.7 points.
The experiments also show that the difference module enhances the BEV feature representation, improving the VPQ and IoU scores by 0.3 points in the long-range evaluation, and by 0.5 and 0.1 points in the short-range evaluation, gains that are confirmed by the repeated runs except for the short-range IoU.
Both evaluations were performed using the Triplet-TST method.
Since Quadruplet-TSST and Triplet-TST perform comparably, Triplet-TST is selected as the default configuration, as it requires fewer parameters and has a lower inference time.
These results emphasise the importance of temporal modelling for accurate Instance Prediction, particularly in long-range scenarios where motion dynamics are more challenging.

Table~\ref{tab:n_temp_blocks} shows how changing the number of temporal attention blocks affects the overall performance of the model.
All configurations correspond to the complete BEVPredFormer model, with the only difference being the number of temporal processing blocks used.
In all experiments, the temporal attention pattern is fixed to Triplet-TST to isolate the impact of block depth.
Increasing the number of blocks from one to two results in the largest performance gains, with improvements of 0.6 points in VPQ and 0.4 points in IoU. However, adding more blocks beyond this point does not yield further VPQ gains, indicating that performance saturates beyond two blocks rather than following a monotonic trend.
For example, increasing from two to four blocks slightly reduces VPQ by 0.2 points and leaves IoU unchanged, while introducing nearly 100 additional GFLOPs and 6.3 million parameters. Even with six blocks, IoU only improves by 0.2 points, at the cost of 190 additional GFLOPs and more than doubling the parameter count.
Therefore, using two temporal blocks offers the best balance between prediction performance and computational efficiency.

Table~\ref{tab:backbone_vt} isolates the contribution of the image feature extractor and the view transform, replacing EfficientViT with EfficientNet-B4 and the attention-based projection of BEVFormer with LSS.
With the temporal module fixed, both components contribute to the final performance. The attention-based view transform has the largest impact: replacing LSS with BEVFormer improves VPQ by 3.6 points and IoU by 3.7 points with EfficientViT, and by 2.7 and 1.6 points with EfficientNet-B4.
Likewise, EfficientViT outperforms EfficientNet-B4 with both view transforms, improving VPQ by 2.7 and 1.8 points with BEVFormer and LSS, respectively.
The gains are larger when both components are combined, which suggests that the attention-based projection benefits from the richer features of EfficientViT.
In terms of cost, EfficientViT requires more GFLOPs than EfficientNet-B4 but runs faster on the GPU, while BEVFormer adds a small number of parameters and GFLOPs over LSS but reduces latency, since it avoids building a dense depth volume for each camera.
Therefore, the selected combination provides the best accuracy with the lowest latency among the evaluated variants.

\begin{table}[t]
\caption{Input image size performance impact. Reported metrics for long-range evaluation using the Triplet-TST temporal model. Forward latency measured as in Table~\ref{tab:efficiency}. Best value is marked in \textbf{bold}.}
\label{tab:img_res}
\begin{tabular*}{\columnwidth}{@{\extracolsep\fill}cccc@{}}
\toprule
\textbf{Resolution} & \textbf{VPQ $\uparrow$} & \textbf{IoU $\uparrow$} & \textbf{Latency (ms) $\downarrow$} \\ \midrule
\(224 \times 480\)  & 34.1          & 41.0          & \textbf{57.7} \\
\(448 \times 800\)  & 35.4          & \textbf{43.3} & 83.2 \\
\(640 \times 1600\) & \textbf{35.5} & \textbf{43.3} & 149.5 \\ \bottomrule
\end{tabular*}
\end{table}

Table~\ref{tab:img_res} illustrates the impact of input image resolution on the model's overall performance.
Following common practices in the literature on detection and prediction, we evaluate three resolutions: the full-width resolution of \(640 \times 1600\), obtained by cropping the upper part of the original \(900 \times 1600\) nuScenes images; the half-resolution of \(448 \times 800\), which is used in most experiments; and a low-resolution setting of \(224 \times 480\).
Image resolution plays a critical role in BEVFormer-based architectures due to its impact on the quality of the 3D projection.
Higher-resolution images result in denser projections that cover more BEV cells with richer spatial detail. However, this comes at a substantial computational cost, which has a particularly significant impact on the runtime of the image feature extractor.
In BEVPredFormer, increasing the resolution from the lowest setting to \(448 \times 800\) substantially improves long-range prediction, with VPQ rising 1.3 points and IoU 2.3 points, while increasing latency by 25.5 ms.
However, further increasing the resolution to the full-width \(640 \times 1600\) only provides a marginal VPQ gain of 0.1 points and no IoU improvement, while significantly reducing inference speed, with a total inference time of 149.5 ms. This demonstrates that \(448 \times 800\) is a practical compromise between accuracy and latency.

\subsection{Robustness Analysis}\label{subsec4.5}

Table~\ref{tab:conditions} reports the performance of BEVPredFormer under different scene conditions in the long-range setting.
As nuScenes does not provide official weather or time-of-day annotations, scenes are classified through keyword matching of their descriptions, which is an approximation. Scenes whose description contains ``night'' or ``dark'' are assigned to the night subset, those containing ``rain'', ``rainy'' or ``wet'' to the rain subset, and the remaining scenes to the day subset. The three scenes recorded at night under rain are assigned to the night subset, so the three subsets are disjoint and together cover the full validation set. Since VPQ and IoU are computed by pooling over all samples (Section~\ref{subsec4.2}), the full-validation score is a weighted average of the scores of the three subsets.

\begin{table*}[b!]
\caption{Long-range performance of BEVPredFormer (\(448 \times 800\)) on disjoint subsets of the nuScenes validation set defined by scene condition. Actors/sample and moving vehicles are computed at the present frame over the evaluated ground-truth vehicles; a vehicle is considered moving if its speed exceeds 0.5~m/s.}
\label{tab:conditions}
\begin{tabular*}{\textwidth}{@{\extracolsep\fill}lcccccc@{}}
\toprule
\textbf{Subset} & \textbf{Scenes} & \textbf{Samples} & \textbf{Actors/Sample} & \textbf{Moving (\%)} & \textbf{VPQ $\uparrow$} & \textbf{IoU $\uparrow$} \\ \midrule
Day      & 111 & 3783 & 10.95 & 29.7 & 34.7 & 42.3 \\
Rain     & 24  & 824  & 14.09 & 17.9 & 38.4 & 47.4 \\
Night    & 15  & 512  & 4.51  & 51.3 & 32.0 & 37.3 \\ \midrule
Full val & 150 & 5119 & 10.81 & 28.1 & 35.4 & 43.3 \\ \bottomrule
\end{tabular*}
\end{table*}

Night conditions reduce VPQ and IoU by 2.7 and 5.0 points, respectively, compared with daytime scenes, mainly due to the difficulty of segmenting poorly illuminated vehicles using camera data alone.
In contrast, the rain subset obtains the highest scores, which is mainly explained by its traffic composition rather than by the weather itself: rainy scenes contain more vehicles per sample, a higher vehicle occupancy of the BEV grid and a lower proportion of moving vehicles, which are the hardest to predict (Table~\ref{tab:speed}). Conversely, night scenes contain fewer vehicles, most of which are moving.

Table~\ref{tab:robustness} evaluates the robustness of BEVPredFormer to errors in camera calibration and ego-pose estimation, the former of which is used for BEV projection and the latter for ego-motion compensation.
Calibration noise is simulated by perturbing the camera extrinsics with a rotation of magnitude \(\theta\) about a random axis, as well as a translation of magnitude \(d\). These are sampled independently for each camera and kept constant over the input sequence. Ego-pose noise is simulated by perturbing the pose of each past frame with a planar translation of magnitude \(d\) and a yaw error \(\psi\), both of which are sampled independently for each frame.
Small errors have a limited impact on both types of noise, whereas larger errors result in clear degradation. In both cases, the rotation component is responsible for most of the performance drop, while translation errors have a minimal effect. BEVPredFormer is more sensitive to calibration errors than to ego-pose errors since its attention-based projection depends directly on the orientation of each camera and angular errors in the projection increase with distance from the camera.
These results suggest that achieving accurate rotation calibration and heading estimation is essential for deploying the model in real-world scenarios.

\begin{table}[t]
\caption{Robustness of BEVPredFormer (\(448 \times 800\)) to calibration noise and ego-pose noise in the long-range setting. Values are the changes in VPQ and IoU with respect to the unperturbed model on the full validation set.}
\label{tab:robustness}
\begin{tabular*}{\columnwidth}{@{\extracolsep\fill}llcc@{}}
\toprule
\textbf{Setting} & \textbf{Noise} & \textbf{$\Delta$VPQ} & \textbf{$\Delta$IoU} \\ \midrule
\multirow{5}{*}{Calibration} & \(\theta = 0.5^\circ\), \(d = 2\) cm  & $-$0.7 & $-$0.3 \\
                             & \(\theta = 1^\circ\), \(d = 5\) cm    & $-$2.4 & $-$1.3 \\
                             & \(\theta = 2^\circ\), \(d = 10\) cm   & $-$7.2 & $-$4.0 \\
                             & \(\theta = 2^\circ\), \(d = 0\) cm    & $-$7.1 & $-$4.0 \\
                             & \(\theta = 0^\circ\), \(d = 10\) cm   & $-$0.3 & $-$0.1 \\ \midrule
\multirow{5}{*}{Ego-pose}    & \(d = 0.1\) m, \(\psi = 0.5^\circ\)   & $-$0.4 & $-$0.2 \\
                             & \(d = 0.25\) m, \(\psi = 1^\circ\)    & $-$1.3 & $-$0.9 \\
                             & \(d = 0.5\) m, \(\psi = 2^\circ\)     & $-$4.0 & $-$3.1 \\
                             & \(d = 0.5\) m, \(\psi = 0^\circ\)     & $-$0.7 & $-$0.4 \\
                             & \(d = 0\) m, \(\psi = 2^\circ\)       & $-$3.5 & $-$2.8 \\ \bottomrule
\end{tabular*}
\end{table}

Despite these results, some limitations should be noted. Following the standard evaluation protocol, only vehicles with at least 40\% camera visibility are considered, so heavily occluded objects, which are particularly relevant for safety, are not assessed, and pedestrians are not included. The analysis of adverse conditions is also limited by the small number of night scenes in nuScenes and the lack of official weather and time-of-day annotations, and the subsets differ in traffic composition, so their scores reflect both the scene condition and the scene content, and rare scenarios, such as unusual manoeuvres or atypical vehicles, are under-represented. Finally, all experiments are conducted on nuScenes, so the generalisation of the model and of the compared methods to other sensor configurations and geographic domains remains to be validated through cross-dataset experiments, which we leave for future work.

\section{Conclusions}\label{sec5}

In this work, we introduce BEVPredFormer, a novel architecture for future Instance Prediction in BEV space.
Our method uses temporal attention modules and multi-scale decoding alongside a two-stage training strategy to efficiently fuse past BEV features and make accurate instance-level predictions over time.
By adapting modules such as PredFormer to operate on BEV difference-guided feature maps, we can capture the spatio-temporal dynamics of traffic scenes effectively.
On the nuScenes dataset, BEVPredFormer matches or surpasses existing methods in VPQ and IoU, particularly in long-range scenarios, at a moderate computational cost.
Qualitative results also show that it accurately predicts vehicle motion in a variety of complex traffic situations.

We conducted extensive ablation studies to explore the impact of different types of temporal attention, the number of attention blocks and the resolution of the input image.
The results show that even lightweight temporal processing improves performance, particularly for long-range predictions, and repeated runs with matched random seeds confirm that both the temporal module and the difference module provide statistically significant improvements.
Using two Triplet-TST attention blocks strikes the best balance between performance and computational efficiency.
Furthermore, we found that increasing the input resolution improves segmentation quality, but this comes at the expense of inference time.
Both the EfficientViT feature extractor and the attention-based view transform contribute to the final performance, with the latter having the largest impact, while also reducing latency compared with LSS.
Overall, these results show that combining an attention-based BEV projection with attention-based temporal processing of BEV features is an effective approach for Instance Prediction.

Nevertheless, this work has some limitations that open several directions for future research.
BEVPredFormer primarily fails when it encounters distant, partially observed, or slow-moving vehicles. Since it produces a single, deterministic future, it cannot represent multiple plausible futures in ambiguous scenarios.
Its performance decreases at night and when there is a large amount of camera rotation or heading error. Additionally, its computational cost is higher than that of lightweight LSS-based methods.
In addition, following the standard protocol for this task, the evaluation is limited to the nuScenes validation set and to vehicles with at least 40\% camera visibility, while adverse conditions and rare scenarios are under-represented in the dataset.
Future work will therefore focus on extending BEVPredFormer to multimodal prediction, evaluating it on other datasets with different sensor configurations and geographic domains, and including pedestrians and heavily occluded objects.
Additionally, we plan to explore fusing camera features with LiDAR or RADAR data to reduce the depth ambiguity of camera-only BEV projection at long range.

\section*{Data and Code Availability}
All experiments were performed using the publicly available nuScenes dataset. The code, configuration files and pre-trained weights for both ranges are publicly available at \url{https://github.com/miguelag99/BEVPredFormer}.

\section*{Publication Note}
\newcommand{\sensorsdoi}{10.3390/s26227206}% TODO: replace with the DOI of the Sensors article
This is the preprint version of an article published in \emph{Sensors} (MDPI) in 2026. Please cite the published version: M.~Antunes-García, S.~Montiel-Marín, F.~Sánchez-García, R.~Gutiérrez-Moreno, R.~Barea, and L.~M.~Bergasa, ``BEVPredFormer: Spatio-Temporal Attention for BEV Instance Prediction in Autonomous Driving,'' \emph{Sensors}, 2026, doi: \href{https://doi.org/\sensorsdoi}{\sensorsdoi}.

\bibliographystyle{IEEEtran}
\bibliography{bibliography}

\end{document}